\documentclass[journal]{IEEEtran}
\usepackage[caption=false]{subfig}
\usepackage{multirow}
\usepackage{amsmath}
\usepackage{amssymb}
\usepackage{lipsum}
\usepackage{cuted}
\usepackage{picinpar, float, graphicx}
\usepackage{booktabs}
\usepackage{array}
\usepackage{longtable} 
\usepackage{epstopdf}
\usepackage{setspace} 
\usepackage{xcolor}
\usepackage{subfig}
\usepackage{gensymb}% for \degree
\usepackage{soul} %strikethrough
\usepackage[normalem]{ulem}
\usepackage{bm}
\usepackage{graphicx}
\usepackage{mathrsfs} 
\usepackage{algorithm}
\usepackage{algorithmic}
\usepackage{threeparttable} % 
\usepackage{multirow}

\usepackage{textcomp,gensymb}
\usepackage{tikz,xcolor,hyperref}

\definecolor{lime}{HTML}{A6CE39}
\DeclareRobustCommand{\orcidicon}{
\begin{tikzpicture}
\draw[lime, fill=lime] (0,0)
circle[radius=0.16]
node[white]{{\fontfamily{qag}\selectfont \tiny \.{I}D}}; 
\end{tikzpicture}
\hspace{-2mm}
}
\foreach \x in {A, ..., Z}{%
\expandafter\xdef\csname orcid\x\endcsname{\noexpand\href{https://orcid.org/\csname orcidauthor\x\endcsname}{\noexpand\orcidicon}}
}
\begin{document}

\title{Capstan-driven Continuum Surgical Robot: Design, Modeling, and Perception}

\author{ Gang Zhang\orcidB{},Yufu Qiu\orcidD{},Junyan Yan\orcidA{}, Wenhui Zeng\orcidC{}, Wenlong Lu\orcidE{} and Shing Shin Cheng\orcidF{} 
\thanks{This work was supported in part by Research Grants Council (RGC) of Hong Kong (CUHK 14217822, CUHK 14207823, CUHK 14211425, and AoE/E-407/24-N), in part by Innovation and Technology Commission of Hong Kong (MHP/096/22, ITS/224/23, ITS/225/23, ITS/353/24, and Multi-scale Medical Robotics Center (InnoHK initiative)), and in part by Curve Robotics Limited, Hong Kong. (Corresponding author: Shing Shin Cheng)

\  Gang Zhang, Yufu Qiu and Junyan Yan are with the Department of Mechanical and Automation Engineering and T Stone Robotics Institute, The Chinese University of Hong Kong, Hong Kong (Email: gangzhang@link.cuhk.edu.hk, junyanyan@mae.cuhk.edu.hk, Yufuqiu@mae.cuhk.edu.hk).

\ Wenhui Zeng is with the School of Mechanical and Electronic Engineering, Wuhan University of Technology, Wuhan 430070, China (Email: wenhuizeng@whut.edu.cn).

\ Wenlong Lu is with state Key Laboratory of Intelligent Manufacturing Equipment and Technology, Huazhong University of Science and Technology (HUST), Wuhan, China (Email: hustwenlong@hust.edu.cn).

\ Shing Shin Cheng is with the Department of Mechanical and Automation Engineering and T Stone Robotics Institute, The Chinese University of Hong Kong, Hong Kong, and also with the Shun Hing Institute of Advanced Engineering and Multi-Scale Medical Robotics Center, Hong Kong (e-mail: sscheng@cuhk.edu.hk).}
\thanks{Manuscript received XX XX, XXXX; revised XX XX, XXXX.}
}
% The paper headers
\markboth{IEEE Transactions on ,~Vol.~X, No.~X, August~2026}%
{Shell \MakeLowercase{\textit{et al.}}: A Sample Article Using IEEEtran.cls for IEEE Journals}

\maketitle
\begin{abstract}
Shape and force sensing have long been critical bottlenecks in the development of compact capstan-driven continuum surgical robots, primarily due to the difficulty of obtaining cable tension information within the confined capstan assembly. To overcome these challenges, this paper presents an integrated design–modeling–sensing approach based on the concept of actuation–perception co-design. A compliant element is introduced into the motor mounting bracket of the drive system, enabling micro-deformation under the cable reaction force and thereby allowing real-time cable tension measurement without occupying the compact capstan space. To address the modeling complexity arising from unconventional joint configurations introduced by the spatial cable routing strategy, a parallel computation framework based on a multibody short-thick-beam model is proposed, which captures shear effects in short beam segments and synergistic multi-cable interactions while achieving real-time performance. Building on this framework, stable shape and force sensing is achieved by incorporating a proximal multi-axis force/torque sensor as an additional measurement anchor. Following this design–modeling–sensing framework, capstan-driven continuum scurgical robots with single- and dual-segment configurations are developed. Experimental results validate the proposed framework in both single- and dual-segment continuum robots, demonstrating real-time tip pose estimation together with contact force and location perception. By enabling cable tension feedback without compromising the compact capstan architecture, the proposed framework makes integrated perception feasible for capstan-driven continuum surgical robots.
\end{abstract}


\section{Introduction}

Continuum robots have been widely adopted in the medical field due to their intrinsic flexibility and ability to navigate deep and tortuous anatomical pathways for minimally invasive interventions. After nearly two decades of development, significant progress has been made in their mechanical design~\cite{seleem2023recent,li2025design,qian2025two}, modeling~\cite{liu2025data, wu2026vector,zhang2026general}, sensing~\cite{gao2024body, zeng20266d}, and control~\cite{zhang2022survey}. The emerging challenge now lies in integrating these advances into a unified framework that enhances robotic intelligence and enables more autonomous surgical operations. In surgical applications, shape sensing is essential for safe navigation through tortuous anatomical pathways, while force sensing is critical for regulating tool-tissue interaction and preventing inadvertent tissue damage. Achieving both capabilities within a compact, clinically deployable instrument therefore represents a key prerequisite for autonomous and intelligent surgical operation.

From a technical perspective, the evolution of continuum robots can be viewed as a layered structure: mechanical design forms the foundation, modeling and sensing constitute the middle layer, and control lies at the top. Each layer is intrinsically linked to the actuation mechanism that defines the robot's architecture and overall performance. Among various mechanical implementations, cable-driven actuation remains the most widely adopted in continuum surgical robots due to its compactness and high transmission efficiency~\cite{dupont2022continuum}.

A stable and reliable design is fundamental to achieving perception and control within this three-layer framework. Regarding actuation, two typical implementations are lead screw drive systems and capstan drive systems~\cite{li2025compact,shao2026deep,yang2026lightweight}. Lead screw drive systems are commonly employed in research platforms because they allow direct coupling with tension sensors, enabling precise measurement of cable forces. For instance, Xiang et al.~\cite{xiang2023learning} used multiple linear motors as actuators for a continuum robot, with miniature force sensors installed at the connection between the linear motors and tendons; the collected data, combined with LSTM, enabled tip force sensing. Yang et al.~\cite{yang2024novel} developed a bearing-based continuum robot system that uses a lead screw drive with force sensors to collect cable tension signals and reconstructs shape through a static model. Similarly, Du et al.~\cite{du2024sensor} utilized cable tension information and a static model of a screw-driven notched continuum robot to obtain the robot's shape and end-effector contact force. Furthermore, Zhang et al.~\cite{zhang2026cable} achieved shape and end contact force estimation by combining static models, cable tension information, and additional six-axis force sensor data. Building on this line of work, research in the last two years further demonstrated the feasibility of contact sensing in lead screw–driven continuum surgical robots, including a variable-length continuum endoscope that uses a tension sensor to monitor drive cable tension for tip force sensing~\cite{zhang2024design}, and a continuum robot for endoscopic submucosal dissection (ESD) that employs a force sensor mounted on a lead screw nut to acquire cable tension for lifting force estimation~\cite{zhang2024designesd}.

While lead screw drives facilitate the integration of sensors for shape and force sensing, their bulky structure and limited actuation speed hinder clinical integration and modular design. In contrast, capstan drive systems offer inherent advantages in compactness, rapid instrument exchange, and compatibility with sterilization procedures, making them not only the preferred choice in commercial surgical platforms developed by companies such as Intuitive Surgical, Connostron, and MicroPort, but also an increasingly necessary architecture for compact continuum surgical robots intended for clinical deployment. These designs have inspired the actuation architecture of cable-driven continuum robots.

Ma et al.~\cite{ma2025vibration} designed a handheld system for bone cutting using a capstan-driven notched continuum robot, achieving a material removal rate of up to 1700 mm³/s. Troncoso et al.~\cite{troncoso2022continuum} developed a compact, truly portable drive mechanism for continuum robots using capstan-driven actuation, enabling exploration in industrial environments. Hong et al.~\cite{hong2022two} designed a novel two-segment continuum robot for maxillary sinus surgery, achieving compact spatial arrangement and rapid instrument exchange through a capstan-driven mechanism. Similar capstan-driven designs have also been explored in the context of continuum surgical robots. Zhang et al.~\cite{zhang2020novel} achieved a compact design for a continuum endoscope using capstan drives. Li et al.~\cite{li2025development} developed novel DNA-inspired robotic flexible surgical instruments using a roll-driven mechanism and established a kinematic model based on constant curvature assumptions. Other examples include continuum surgical robots developed by Feng et al.~\cite{feng2022development} for transoral surgery and by Li et al.~\cite{li2023robotic} for transanal endoscopic microsurgery. %These works collectively demonstrate the mechanical feasibility of capstan-driven Continuum surgical robots, yet real-time sensing and accurate modeling remain open challenges in this configuration, which motivates the design framework presented in this paper.

\subsection{Challenge and Contributions}
However, the capstan drive design choice introduces a fundamental limitation: the compact capstan assembly makes it difficult to integrate tension sensors, leaving cable tension unmeasurable and thereby preventing the realization of intrinsic shape and force sensing. To accommodate multiple cables within a small-diameter instrument body, a spatial cable routing strategy is necessary, which not only constrains the available space in the central working channel but also introduces unconventional joint configurations in which cables interact with beam elements at non-planar contact points. This, in turn, compounds the modeling challenge: while existing approaches commonly treat the flexible sections as slender beams based on Euler–Bernoulli or Cosserat beam theories, the flexible segments in notched continuum robots are short and thick relative to their cross-section, rendering slender-beam assumptions inaccurate and causing these models to underestimate shear deformation. The non-planar cable-beam interactions further introduce coupling effects that standard beam models are not designed to capture. Accurately characterizing these effects is therefore essential for achieving reliable shape and force estimation in capstan-driven systems.

To address these challenges, this paper proposes a novel integrated actuation–perception design scheme for notched continuum robots, enabling perception through the combination of mechanical design and a modeling framework based on short beams. The contributions of this paper are as follows:
\begin{enumerate}
\item A mechanical design scheme that integrates perception and actuation is proposed for capstan-driven continuum robots, enabling real-time cable tension monitoring without occupying the compact capstan space. 
\begin{itemize}
\item A compliant element is introduced into the motor mounting bracket, amplifying the micro-deformation induced by the cable reaction force. Cable tension is obtained by mapping the measured strain gauge deformation to the corresponding tension value.
\item The sensing signal is acquired at up to 1000 Hz, determined by the data acquisition hardware. Experiments demonstrate that over a measurement range of 0-9.5 N, the average tension estimation error is 0.12 N and the maximum error does not exceed 0.4 N. 
\end{itemize}

\item A compact spatial cable arrangement is developed to accommodate eight cables within a two-segment continuum robot of 3.5\,mm diameter and less than 0.4\,mm wall thickness. By distributing the cables along spatially staggered paths, the maximum number of cable channels required in each rigid ring is reduced from eight to four, enabling the multi-cable arrangement within the limited radial space.

\item  A short-thick-beam static model is developed to address the unconventional joint configurations arising from spatially interlaced cable routing in capstan-driven continuum robots, enabling real-time shape and contact force sensing.
\begin{itemize}
\item Unlike conventional slender-beam models based on Euler--Bernoulli or Cosserat beam theories, the proposed model explicitly captures shear deformation in short, thick beam segments and the synergistic loading effects of multiple cables under non-planar contact geometry.
\item A group-parallel computation strategy is adopted in which the mechanical variables of all beam elements are organized as a batch matrix operation, avoiding element-by-element iteration. This enables model update rates exceeding 200\,Hz, representing a more than tenfold improvement over the non-parallelized variant.
\item Combined with the proximal sensing strategy in~\cite{zhang2026cable}, the developed system enables shape and contact perception in both single- and dual-segment configurations.
\end{itemize}
\end{enumerate}

The remainder of this paper is organized as follows. Section II describes the design, static modeling, and perception of the proposed system. Section III presents the experimental setup and results, including tension calibration, model validation, and contact force sensing. Section IV discusses the results and limitations. Section V concludes the paper.

 \begin{figure*}[t]
\centering  
\includegraphics[width=180mm]{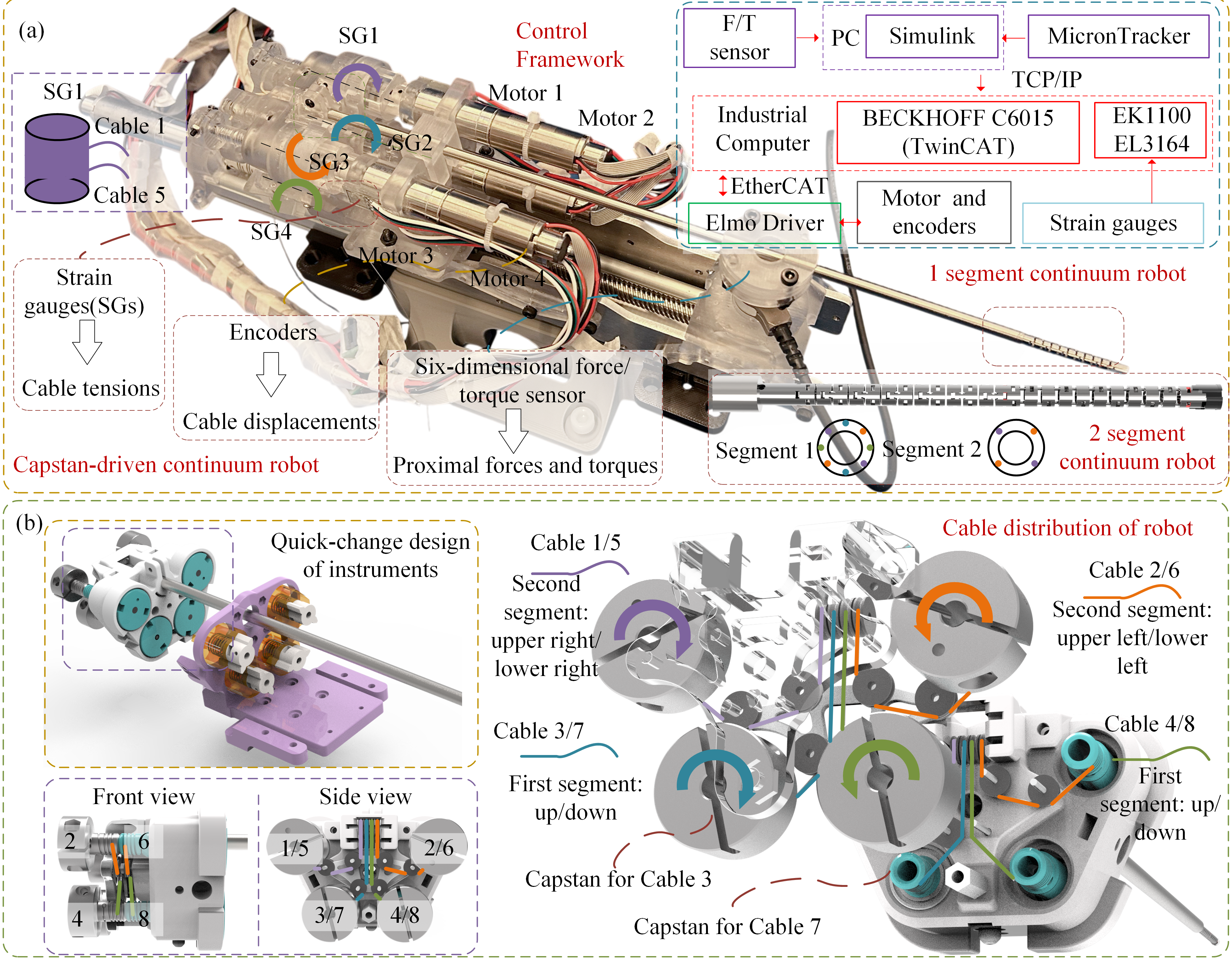}
\caption{Design and principle of a capstan-driven continuum surgical robot. (a) The basic components of the robot: mechanical design and control framework hardware. (b) The distribution of cables in the equipment and the principle of cable drive.
}
%{\color{blue} better to use the same font size in the diagram.}
\label{Struc}   
\vspace{-15 pt}
\end{figure*}

\section{Methods} 

%This section presents the proposed capstan-driven continuum surgical robot system from three aspects: mechanical design, static modeling, and the shape and force sensing framework.

\subsection{Structure} 
Based on the requirements for rapid instrument exchange, a capstan-driven surgical robot system was designed. The drive system is compatible with the developed continuum surgical robots, meeting the basic demands of surgical operation. The overall robot system is shown in Fig.~\ref{Struc}(a). The mechanical hardware comprises a drive unit and the corresponding continuum surgical robots. The drive unit consists of four motors (DCX14, Maxon, Switzerland), each mounted on a dedicated motor bracket and equipped with a capstan for torque transmission.
Each motor bracket is designed with thin-walled beam structures on both lateral faces. Under the reaction force generated by cable tension on the motor, these thin-walled beams undergo slight but measurable elastic deformation. A strain gauge (BF1K-3EB, RuneKee, China) is bonded to each bracket face to capture this deformation, and the measured signal is mapped to the corresponding cable tension through a calibrated neural network.

The electrical system is organized into three layers: a host computer, a slave computer, and an execution unit. The host computer is a personal computer (GTX\,1650-i5, HP, USA) running MATLAB Simulink, which communicates with the slave computer via TCP/IP. Data from the tracking camera (MicronTracker, Claron Technology, Canada), and six-axis force/torque sensor (Nano 17, ATI, USA, range: Fx and Fy:12 N, Fz:17 N, Tx, Ty, Tz: 120 N·mm, resolution: 1/320 N for force and 1/64 N·mm for torque, sampling rate: 7000 Hz) are directly imported into the Simulink platform. The core of the slave control unit is a Beckhoff industrial computer (C6015, Beckhoff, Germany) equipped with a data acquisition module (EL3164, Beckhoff, Germany) for acquiring strain gauge signals, communicating with the execution unit in real time via the EtherCAT protocol. The execution unit includes four servo drivers (Gold Solo Twitter, Elmo, Denmark) and four Maxon motors.

The continuum surgical robot section can drive up to two segments of the continuum robot; here, a two-segment configuration is taken as a representative example. Four docking capstans interface with the corresponding capstans on the drive unit. Each docking capstan features two winding posts-an upper and a lower-both secured to an optical axis with bolts, with the lower post integrated directly into the capstan body. Two sets of cables (diameter: 0.27 mm, TESAC, Japan), namely $C_1$-$C_5$, $C_2$-$C_6$, $C_3$-$C_7$, and $C_4$-$C_8$, are connected to the upper and lower winding posts, respectively. Each cable pair corresponds to one degree of freedom of a single segment, and all eight cables are routed to the continuum robot section via rollers.

\begin{figure*}[t]
\centering  
\includegraphics[width=180mm]{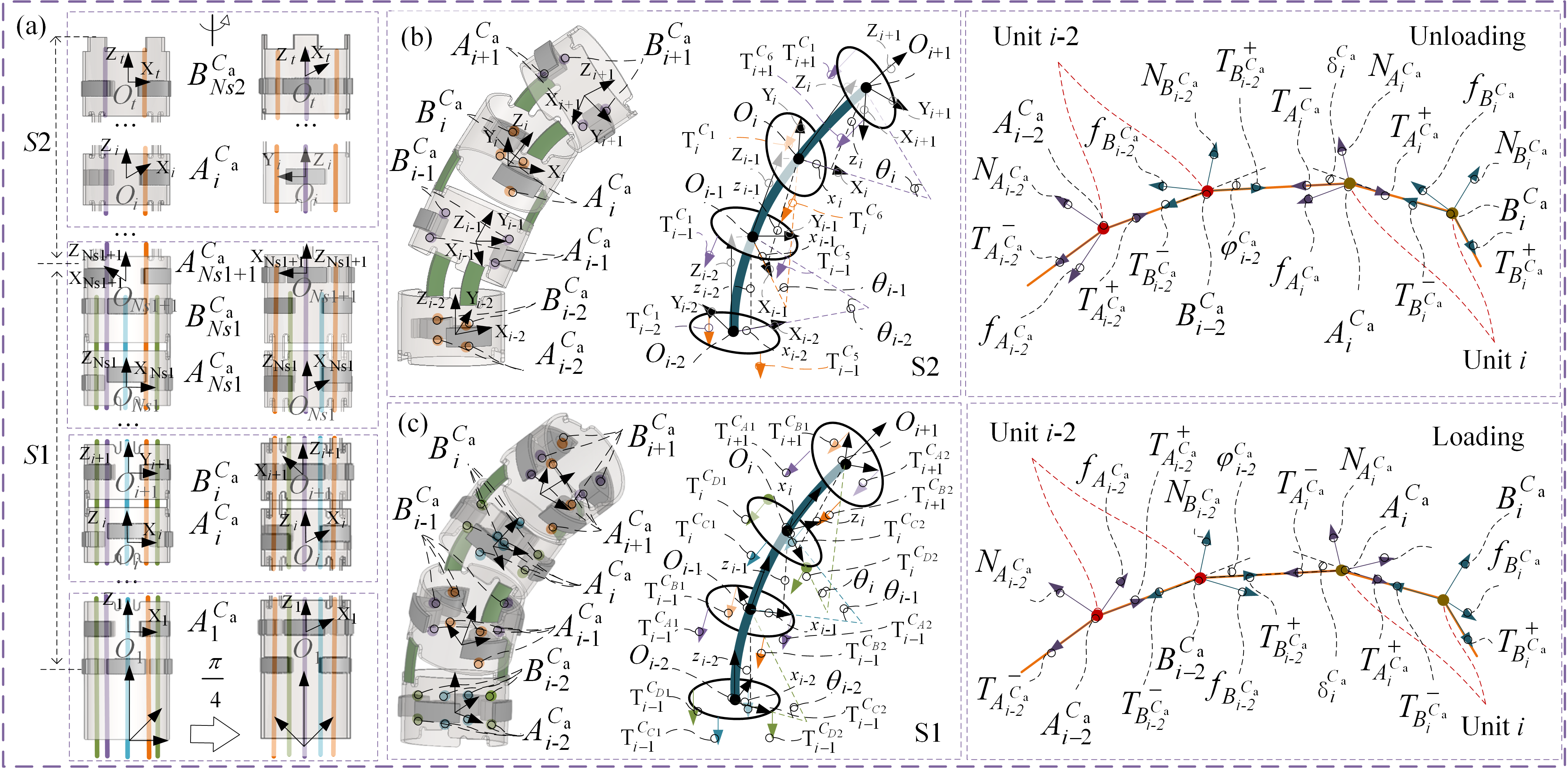}
\caption{Design of continuum surgical robot and internal tension variations. (a) Body design of  continuum surgical robot, (b) Spatial layout of cables, (c) Tension transmission on cables. 
}
\label{CR}   
\vspace{-10 pt}
\end{figure*}

The design of the continuum robot is shown in Fig.~\ref{CR}(a). The flexible structure is formed by laser-cutting slots into a Nitinol alloy tube, producing alternating flexible beam segments and rigid rings. Multiple square slots are machined into the rigid rings to accommodate cable routing components. The continuum robot adopts an intersecting beam configuration: beams within the same rigid ring form a group, with adjacent groups oriented 90° apart to enable bidirectional bending. The beams in the second segment are rotated 45° about the axial direction relative to those in the first segment. Unlike conventional designs in which all cables pass through the same rigid ring, the cables are here divided into two groups and routed through adjacent rings, enabling a compact multi-cable layout under small-diameter constraints. As a result, each rigid ring requires a maximum of four cable channels for an eight-cable-driven continuum robot.

\subsection{Model} 

Define the homogeneous transformation matrix for translation and rotation. The exponential map of a twist $\boldsymbol{\xi} = (\boldsymbol{\omega}, \mathbf{v})^{\mathrm{T}} \in \mathbb{R}^6$ is given by:
\begin{equation}
\mathbf{T}(\boldsymbol{\xi}, \theta) = e^{\hat{\boldsymbol{\xi}}\theta}, \quad 
\hat{\boldsymbol{\xi}} = \begin{bmatrix} \hat{\boldsymbol{\omega}} & \mathbf{v} \\ \mathbf{0} & 0 \end{bmatrix},
\end{equation}
where $\hat{\boldsymbol{\omega}}$ is the skew-symmetric matrix of $\boldsymbol{\omega}$. Elementary twists used in this work are:

\begin{itemize}
\small
\item Translation along $z$: $\boldsymbol{\xi}_{\text{tz}} = (0,0,0;0,0,1)^{\mathrm{T}}$, $e^{\hat{\boldsymbol{\xi}}_{\text{tz}} d} = \mathbf{T}(0,0,d)$.
\item Translation along $x$: $\boldsymbol{\xi}_{\text{tx}} = (0,0,0;1,0,0)^{\mathrm{T}}$, $e^{\hat{\boldsymbol{\xi}}_{\text{tx}} d} = \mathbf{T}(d,0,0)$.
\item Rotation about $y$: $\boldsymbol{\xi}_{\text{ry}} = (0,1,0;0,0,0)^{\mathrm{T}}$, $e^{\hat{\boldsymbol{\xi}}_{\text{ry}} \theta} = \mathbf{R}_y(\theta)$.
\item Rotation about $z$: $\boldsymbol{\xi}_{\text{rz}} = (0,0,1;0,0,0)^{\mathrm{T}}$, $e^{\hat{\boldsymbol{\xi}}_{\text{rz}} \theta} = \mathbf{R}_z(\theta)$.
\end{itemize}

The robot's centerline is divided into alternating flexible deformable parts (blue line) and rigid parts (black disk). The transfer matrix for the bottom node of the beam of the first segment on the centerline is defined as ${T}_{b,i}^{(1)}$, and the transfer matrix for the tip node as ${T}_{t,i}^{(1)}$.
 Let $\mathbf{T}_{t,0}^{(1)} = \mathbf{I}_4$. For $i = 1,\dots,12$,

\begin{equation}
\left\{ \begin{array}{l}
\mathbf{T}_{b,i}^{(1)}= \mathbf{T}_{t,i-1}^{(1)} \cdot e^{\hat{\boldsymbol{\xi}}_{\text{tz}} s_1}, \\
\mathbf{T}_{t,i}^{(1)}= \mathbf{T}^{(1)}_{b,i} \cdot e^{\hat{\boldsymbol{\xi}}_{\text{tx}} x_i} \cdot e^{\hat{\boldsymbol{\xi}}_{\text{tz}} z_i} \cdot e^{\hat{\boldsymbol{\xi}}_{\text{ry}} \theta_i},
\end{array} \right.
\label{Ts1_1}
\end{equation}
where $i$ is the beam number, $x_i$, $z_i$ are the position of the beam tip, $\theta_i$ is the deflection angle at the tip of beam $i$. $s_1$ is the distance between the beams in segment 1.
The coordinates of the beam base and tip are extracted as:

\begin{equation}
\left\{ \begin{array}{l}
{\bf{P}}_{b,i} = {\bf{T}}_{b,i}(1:3,4),\\
 {\bf{P}}_{t,i} = {\bf{T}}_{t,i}(1:3,4).
\end{array} \right.
\label{P_S1_1}
\end{equation}

The initial pose of the second section is:

\begin{equation}
\mathbf{T}_{t,0}^{(2)} = \mathbf{T}_{t,12}^{(1)} \cdot e^{\hat{\boldsymbol{\xi}}_{\text{tz}} s_{{e1}}} \cdot e^{\hat{\boldsymbol{\xi}}_{\text{rz}} (-\pi/4)},
\end{equation}
where $s_{e1}$ is the distance between the end of the last beam in the first segment and the base of the first beam in the second segment.
Similarly, the transfer matrix ${T}_{b,i}^{(2)}$ of the bottom node and the transfer matrix ${T}_{t,i}^{(2)}$ of the top node of the beams in segment 2 can be calculated using the following equations.
\begin{equation}
\left\{ \begin{array}{l}
\bf{T}_{b,i}^{(2)}= \bf{T}_{t,i-1}^{(2)} \cdot e^{\hat{\boldsymbol{\xi}}_{\text{tz}} s_2}, \\
\bf{T}_{t,i}^{(2)}= \bf{T}_{b,i}^{(2)} \cdot e^{\hat{\boldsymbol{\xi}}_{\text{tx}} x_i} \cdot e^{\hat{\boldsymbol{\xi}}_{\text{tz}} z_i} \cdot e^{\hat{\boldsymbol{\xi}}_{\text{ry}} \theta_i},
\end{array} \right.
\end{equation}
where $s_2$ is the distance between the beams in the second segment.
Eight cables $C_a$ ($a=1,\dots,8$) are evenly distributed. 
For the first section, lower contact points (for cables $C_1,C_3,C_5,C_7$) are:
\begin{equation}
\mathbf{T}_{C_a,b,j} = \mathbf{T}_{t,i} \cdot e^{\hat{\boldsymbol{\xi}}_{\text{tz}} d_1} \cdot e^{\hat{\boldsymbol{\xi}}_{\text{rz}} \varphi_a} \cdot e^{\hat{\boldsymbol{\xi}}_{\text{ry}} (\pi/2)} \cdot e^{\hat{\boldsymbol{\xi}}_{\text{tx}} (d_c/2)},
\end{equation}
where $i$ is an even number, $d_1$ is the distance from the tip of the beam to the bottom of the cable channel, $d_c$ is the diameter of the robot, and $\varphi_a$ are the circumferential angles:
\begin{equation}
\varphi_a \in \left\{0,\ {\pi}/{4},\ {\pi}/{2},\ {3\pi}/{4},\ \pi,5\pi/{4},\ {3\pi}/{2},\ {7\pi}/{4}\right\}.
\end{equation}

Upper contact points (for $C_2,C_4,C_6,C_8$) are:
\begin{equation}
\mathbf{T}_{C_a,b,i} = \mathbf{T}_{t,i} \cdot e^{\hat{\boldsymbol{\xi}}_{\text{tz}} d_1} \cdot e^{\hat{\boldsymbol{\xi}}_{\text{rz}} \varphi_a} \cdot e^{\hat{\boldsymbol{\xi}}_{\text{ry}} (\pi/2)} \cdot e^{\hat{\boldsymbol{\xi}}_{\text{tx}} (d_c/2)},
\end{equation}
where $i$ is an odd number, then translated to the channel tip:
\begin{equation}
\mathbf{T}_{C_a,t,i} = \mathbf{T}_{C_a,b,i} \cdot e^{\hat{\boldsymbol{\xi}}_{\text{tx}} (D_c)}.
\label{T_Ca}
\end{equation}
where $D_c$ is the height of the channel. Similarly, we can use Eq~\eqref{P_Ca} to obtain the spatial positions of the bottom and top points of the channel. 
\begin{equation}
\left\{ \begin{array}{l}
{\bf{P}}_{C_a,2j} = {\bf{T}}_{C_a,b,i}(1:3,4),\\
{\bf{P}}_{C_a,2j+1} = {\bf{T}}_{C_a,t,i}(1:3,4).
\end{array} \right.
\label{P_Ca}
\end{equation}

Using the same method, we can obtain the positions of channels in the second segment.
For a given cable, order points along the path and compute vectors:
\begin{equation}
\mathbf{v}_j = \mathbf{P}_{C_a,j+1} - \mathbf{P}_{C_a,j}.
\end{equation}

The deflection angle at node $j$ is:
\begin{equation}
\Delta\phi_{c,j} = \arccos\left( \frac{\mathbf{v}_j \cdot \mathbf{v}_{j+1}}{\|\mathbf{v}_j\|\,\|\mathbf{v}_{j+1}\|} \right).
\label{delta}
\end{equation}

Based on the Capstan friction model~\cite{du2024sensor}, the tension in cable $C_a$ for beam $i$ is:
\begin{equation}
\left\{ \begin{array}{l}
T_{A^{{C_a}}_i}^{-} = T_{C_a,0} \cdot e^{ \left( -\mu \sum_{j=1}^{i_j} \Delta\phi_{c,j} \right)}, Loading,\\
T_{A^{{C_a}}_i}^{-} = T_{C_a,0} /e^{ \left( -\mu \sum_{j=1}^{i_j} \Delta\phi_{c,j} \right)},Unloading,
\label{T_trans}
\end{array} \right.
\end{equation}
where, for cables 1/3/5/7, when $i$ is odd, $i_j = i$, and when $i$ is even, $i_j = i - 1$. For cables 2/4/6/8, when $i$ is odd, $i_j = i$, and when $i$ is even, $i_j = i + 1$.
The vector form of the force exerted by the cable $C_a$ on beam element $i$ is:
\begin{equation}
\mathbf{F}_{C_a,i} = T_{A^{{C_a}}_i}^{-} \frac{\mathbf{v}_i}{\|\mathbf{v}_i\|} ,
\end{equation}

The total force acting on the tip of the $i$-th beam element (in the global frame) is the sum over all eight cables:
\begin{equation}
\mathbf{F}_i^{\text{global}} = \sum_{a}^{8} \mathbf{F}_{C_a,io}+\mathbf{F}_c,
\label{F_glo}
\end{equation}
where $\mathbf{F}_c$ is the contact force. For cables 1, 3, 5, and 7, when $i$ is odd, $i_o = i$; when $i$ is even, $i_o = i + 1$. For cables 2, 4, 6, and 8, when $i$ is odd, $i_o = i$; when $i$ is even, $i_o = i - 1$.
When calculating the deformation of the beam in the local beam base coordinate system, the force needs to be transformed into the local coordinate system.
The rotation matrix $\mathbf{R}_i$ from the global frame $\{O_{A_0}\}$ to the local frame $\{O_{A_i}\}$ attached to the $i$-th beam element tip can be represented by:
\begin{equation}
\mathbf{R}_i = \mathbf{R}_{i-1} \cdot 
\begin{cases}
e^{\hat{\boldsymbol{\xi}}_{\text{ry}} \theta_{i-1}}, & \text{if } i \text{ is odd},\\
e^{\hat{\boldsymbol{\xi}}_{\text{rx}} \theta_{i-1}}, & \text{if } i \text{ is even},
\end{cases}
\label{eq:R_accum}
\end{equation}
where $\theta_{i-1}$ is the bending angle of the $(i-1)$-th beam element obtained from the beam deformation model. 
For the second section of the robot, an additional initial rotation about $z$ by $-\pi/4$ is applied:
\begin{equation}
\mathbf{R}_{N_{s1}+1} = \mathbf{R}_{N_{s1}} \cdot e^{\hat{\boldsymbol{\xi}}_{\text{rz}} (-\pi/4)} \cdot e^{\hat{\boldsymbol{\xi}}_{\text{ry}} \theta_{N_{s1}}},
\end{equation}
where $N_{s1}$ is the number of beams in the first segment. The same force expressed in the local frame $\{O_{A_i}\}$ is
\begin{equation}
\mathbf{F}_{i}^{\text{local}} = \mathbf{R}_i^{-1} \,\mathbf{F}_{i}^{\text{global}}.
\label{F_loc}
\end{equation}

The radial force $F_{r,i}$ and axial force $F_{z,i}$ used in the beam deformation model are extracted according to the rule:
\begin{equation}
F_{r,i} = 
\begin{cases}
F_{x,i}^{\text{local}}, & \text{if } i \text{ is odd}\\[2pt]
F_{y,i}^{\text{local}}, & \text{if } i \text{ is even}

\end{cases}
,
F_{z,i} = F_{z,i}^{\text{local}}.
\label{F_all}
\end{equation}

The moment about the tip of the $i$-th beam element is generated by the cable forces acting through moment arms. The forces and lever arms on the cable are transformed into a local coordinate system. The total moment from cables can be calculated by:
\begin{equation}
\left\{ \begin{array}{l}
\mathbf{M}_i^{\text{cables}} = \sum_{a=1}^{8} \left( \mathbf{r}_{C_a,i}^{local} \times \mathbf{F}_{C_a,i}^{\text{local}}  \right),\\
\mathbf{r}_{C_a,i}^{local} =\mathbf{R}_i^{-1}(\mathbf{P}_{C_a,mj}- \mathbf{P}_{\text{t},i}),\\
\mathbf{F}_{C_a,i}^{\text{local}} = \mathbf{R}_i^{-1} \,\mathbf{F}_{C_a,i},
\end{array} \right.
\label{M_loc}
\end{equation}
where $\mathbf{F}_{C_a,i}^{\text{local}}$ and $\mathbf{r}_{C_a,i}^{\text{local}}$ are the lever arm and force exerted by the cable on the tip of beam $i$ in the local coordinate system.
For cables 1, 3, 5, and 7, when $i$ is odd, $mj = i+2$; when $i$ is even, $mj = i + 3$. For cables 2, 4, 6, and 8, when $i$ is odd, $mj = i+2$; when $i$ is even, $mj = i +2$.
For contact force $\mathbf{F}_c$ acts at point $\mathbf{P}_{F_c}$, its contribution is
\begin{equation}
\left\{ \begin{array}{l}
\mathbf{M}_i^{F_c} = \mathbf{r}_{F_c,i}^{local} \times (\mathbf{R}_i^{-1} \mathbf{F}_c),\\
\mathbf{r}_{F_c,i}^{local} = \mathbf{R}_i^{-1}(\mathbf{P}_{F_c} - \mathbf{P}_{\text{t},i}).
\end{array} \right.
\label{M_Fc_loc}
\end{equation}

Assume that the contact point is at a distance $s_c$ from the proximal end along the robot backbone and a rotation angle $\theta_c$ about the $z$-axis. ${{\bf{P}}_c}(t)$ can be expressed by Eq.~\eqref{M10},
{\small
\begin{equation}
\left\{ \begin{array}{l}
{{\bf{T}}_{F_c}}(t) ={{\bf{T}}_{N_c}}(t) \cdot e^{\hat{\boldsymbol{\xi}}_{\text{tz}} s_{N_c}} \cdot e^{\hat{\boldsymbol{\xi}}_{\text{ry}} \theta_{c}}\cdot e^{\hat{\boldsymbol{\xi}}_{\text{ry}} \pi/2}{\bf{Trz}}(\frac{d_c}{2}) ,\\
s_{N_c}=s_c-\sum_{i=1}^{Nc-1}(s_m+h_i),\\
\left[ {\begin{array}{*{20}{c}}
{{\bf{R}}_c(t)}&{{\bf{P}}_c(t)}\\
0&1
\end{array}} \right] = {{\bf{T}}_c}(t),
\end{array} \right. 
\label{M10}
\end{equation}}
where $N_c$ is the joint number to which the force is applied.  $h_i$ is the beam height. $s_m$ is the spacing between the beams. For the first segment, $m=1$, and for the second segment, $m=2$. The total moment in the local coordinate system can be calculated by:
\begin{equation}
\mathbf{M}_i = ({\mathbf{M}_i^{\text{cables}} + \mathbf{M}_i^{F_c}})y,
\label{M_all}
\end{equation}
i.e., the $y$-component of the total moment rotated to the local frame.
The three quantities required by the beam deformation model (TBCM) for each beam element $i$ are $F_{r,i}$, $F_{z,i}$ and  $M_i$, calculated via Eqs.~(\ref{F_all}) and (\ref{M_all}).
Here we treat all the beams as an N×3 group to enable parallel computation. The forces and moments at the beam tips are written as an N×3 matrix, where N is the number of beams.
{\small
\begin{equation}
\left\{ \begin{array}{l}
{\bf{r}} =[{r}_{1},{r}_{2},...,{r}_{N}],{\bf{z}} =[{z}_{1},{z}_{2},...,{z}_{N}],\\
{\bf{\Theta}} =[{\theta}_{1},{\theta}_{2},...,{\theta}_{N}],{\bf{F}}_{r} =[{F}_{r,1},{F}_{r,2},...,{F}_{r,N}],\\
{\bf{F}}_{z} =[{F}_{z,1},{F}_{z,2},...,{F}_{z,N}],{\bf{M}} =[{M}_{1},{M}_{2},...,{M}_{N}].
\end{array} \right.
    \label{M_m}
\end{equation}}

These are subsequently used in the dimensionless load definitions:
\begin{equation}
{\bf{f}}_{r} = \frac{{\bf{F}}_{r}\,h^2}{E I},\quad
{\bf{f}}_{z,i} = \frac{{\bf{F}}_{z}\,h^2}{E I},\quad
{\bf{m}}= \frac{{\bf{M}}\,h}{E I},
\end{equation}
where $h$ is the beam height, $I$ the second moment of area, and $E$ the Young's modulus.
The shear influence coefficient can be calculated by:

\begin{equation}
    k_{s,i} = \frac{k_1 (1+\nu) w^2}{6 h^2}, \qquad
    k_1 = \frac{12 + 11\nu}{10 + 10\nu}
    \label{eq:ks}
\end{equation}
where $\nu$ is the Poisson's ratio, $w$ is the thickness of the beam. The effective 2×2 stiffness matrix is expressed as:

\begin{equation}
    \mathbf{K}_i(k_{s,i},{\bf{f}}_z) = \mathbf{G} + {\bf{f}}_{z} \mathbf{P} +{\bf{f}}_{z}^2 \mathbf{Q},
    \label{eq:K_general}
\end{equation}
where the coefficient matrices $\mathbf{G}$, $\mathbf{P}$, $\mathbf{Q}$ depend only on $k_{s,i}$ \cite{li2025chained,chen2015kinetostatic}.
The dimensionless transverse displacement $\hat{\delta}{\bf{r}} = \delta{{\bf{r}}}/h$ and rotation angle $\hat {\mathbf{\Theta}} = {\bf{\Theta}}$ are obtained by solving:

\begin{equation}
    \begin{bmatrix}
        \hat{\delta}{\bf{r}}\\[0.5ex]
        \hat{\mathbf{\Theta}}
    \end{bmatrix}
    = \mathbf{K}_i^{-1}
    \begin{bmatrix}
        {\bf{r}} \\[0.5ex]
        {{\bf{m}}}
    \end{bmatrix}.
    \label{eq:primary_solve}
\end{equation}
Explicitly, using Cramer's rule:

\begin{equation}
\left\{ \begin{array}{l}
\hat{\delta}{\bf{r}} = \frac{K_{22}{\bf{f}}_r - K_{12}{\bf{m}}}{K_{11}K_{22} - K_{12}K_{21}},\\
\hat{\Theta}_i = \frac{-K_{21}{\bf{f}}_r+ K_{11}{\bf{m}}}{K_{11}K_{22} - K_{12}K_{21}}.
\end{array} \right.
\end{equation}

The dimensionless displacement $\hat{\delta}{\bf{z}} = \delta{{\bf{z}}}/h$ is corrected by geometric nonlinearity and higher-order coupling terms:
\begin{equation}
    \hat{\delta}_{\bf{z}} = \frac{w^2 {\bf{f}}_{z}}{12 h^2}
    - \frac{1}{2} \boldsymbol{\xi}_i^\top \mathbf{U} \boldsymbol{\xi}_i
    - {\bf{f}}_{z} \, \boldsymbol{\xi}_i^\top \mathbf{V} \boldsymbol{\xi}_i,
    \label{eq:delta_x_correction}
\end{equation}
with $\boldsymbol{\xi}_i = \begin{bmatrix} \hat{\delta}_{\bf{r}} & \hat{\Theta} \end{bmatrix}^\top$, where $\mathbf{U}$ and $\mathbf{V}$ are coefficient matrices capturing geometric and higher-order effects, respectively. Their explicit expressions can be found in \cite{li2025chained,chen2015kinetostatic}.
Finally, the physical displacements are recovered by:
\begin{equation}
    {\bf{r}} = \hat{\delta}_{\bf{r}} \, h, \quad
    {\bf{z}} = h-\hat{\delta}_{{\bf{z}}} \, h, \quad
    \Theta = \hat{\Theta}.
    \label{beam_end}
\end{equation}

\renewcommand{\algorithmicrequire}{ \textbf{Input:}} 
\renewcommand{\algorithmicensure}{ \textbf{Output:}} 
\begin{algorithm}[ht]
\caption{The solution process of the model.}
\begin{algorithmic}[1]
\REQUIRE{Input forces $T_{C_a,0}$, contact force $\bf{F_c}$, contact distance $s_c$ and rotation angle $\theta_c$.}
\ENSURE{${\bf{P}}_{b,i}$,${\bf{P}}_{t,i}$,${\bf{P}}_{C_a,j}$}
\STATE Initial values: $ {\bf{r}}={\bf{0}}, {\bf{z}}=h,\Theta={\bf{0}},$  number of iterations $n=0$, iteration error $e=1$.
\WHILE{ $e >e_{Limit}$}
\STATE Solve for the position of the beam base and tip ${\bf{P}}_{b,i}$ and ${\bf{P}}_{t,i}$ using Eq.~\eqref{P_S1_1}.
\STATE Solve for the position of the cable channel ${\bf{P}}_{C_a,j} $ using Eq.~\eqref{P_Ca}.
\STATE Solve for cable deflection angle $\Delta\phi_{c,j}$ using Eq.~\eqref{delta}.
\STATE Solving for cable tension $T_{A^{{C_a}}_i}^{-}$ using Eq.~\eqref{T_trans}.
\STATE Solve for the forces $\mathbf{F}_i^{\text{global}} $ at the tips of the beam group. In the global coordinate system using Eq.~\eqref{F_glo}.
\STATE Transform the forces at the tips of the beam group from the global coordinate system to the local coordinate system, $\mathbf{F}_i^{\text{local}} \leftarrow$ Eq.~\eqref{F_loc}.
\STATE Solve for the torques exerted by the cables at the tips of the beam group, $\mathbf{M}_i^{\text{cable}} \leftarrow$ Eq.~\eqref{M_loc}.
\STATE Solve for the moments of external forces acting at the tips of the beam group, $\mathbf{M}^{F_c}_i \leftarrow$ Eq.~\eqref{M_Fc_loc}.
%\STATE $F_{r,i}$ and $F_{z,1}  \leftarrow$ Eq.~\eqref{F_all}, $M_{i}  \leftarrow$ Eq.~\eqref{M_all}.
\STATE Solve for the force group ${\bf{F}}_{r}$, ${\bf{F}}_{z}$, and moment group ${\bf{M}}$ acting on the tip of the beam group on the deformation plane using Eq.~\eqref{M_m}.
\STATE Solve for the deformation of the beam group, $[{\bf{z}}^n,{\bf{r}}^n,{\bf{\Theta}}^n]\leftarrow$ Eq. \eqref{beam_end}.
\STATE Update the initial deformation of the beam group, $[{\bf{z}}^{n+1},{\bf{r}}^{n+1},{\bf{\Theta}}^{n+1}]\leftarrow [{\bf{z}}^n,{\bf{r}}^n,{\bf{\Theta}}^n]$.
\STATE $n=n+1$.
\STATE $e_\theta=max(abs(\Theta^{n}-\Theta^{n-1}))$.
\ENDWHILE
%\STATE $[{\bf{z}},{\bf{r}},{\bf{\Theta}}]\leftarrow [{\bf{z}}^n,{\bf{r}}^n,{\bf{\Theta}}^n]$.
\RETURN {${\bf{P}}_{b,i}$,${\bf{P}}_{t,i}$,${\bf{P}}_{C_a,j}$}.
\end{algorithmic}
\label{Algorithm_F2S}
\end{algorithm}

\subsection{Perception Framework}

Building on this model and following the proximal sensing strategy introduced in~\cite{zhang2026cable}, the perception is formulated as a two-step decoupled estimation problem. As illustrated in Fig.~\ref{Struc}, the sensing modalities comprise three sources: cable tensions $T_{C_a,0}$ ($a=1,\ldots,8$) from strain gauges, cable displacements $\Delta L_{C_a}$ from encoders, and the proximal force $\mathbf{F}_s$ and moment $\mathbf{M}_s$ from the six-axis F/T sensor.

\subsubsection{Problem Formulation}

Let $\mathbf{F}_c\in\mathbb{R}^3$ denote the unknown contact force acting at an unknown backbone location $s_c\in[0,L]$. Under quasi-static equilibrium, the theoretical proximal moment is:
\begin{equation}
    \begin{cases}
    \mathbf{M}_\mathrm{th} =
    \displaystyle\sum_{a=1}^{8}
    \mathbf{r}_{C_a,0}\times\mathbf{F}_{C_a,0}
    + \mathbf{r}_{F_c}(s_c)\times\mathbf{F}_c,\\[8pt]
    \mathbf{r}_{C_a,0} =
    \underbrace{(\mathbf{P}_{t,0}-\mathbf{P}_{sfs})
    }_{\text{axial arm}}
    +\underbrace{
    \tfrac{d_c}{2}
    [-s\phi_a,\,c\phi_a,\,0]^\top
    }_{\text{circumferential offset arm}},
    \end{cases}
    \label{eq:moment_th}
\end{equation}
where $\mathbf{r}_{C_a,0}$ is the lever arm of cable $C_a$ at the proximal entry point relative to the F/T sensor origin $\mathbf{P}_{sfs}$, and $\mathbf{r}_{F_c}(s_c) =\mathbf{P}_{F_c}(s_c) - \mathbf{P}_{sfs}$ is the lever arm of the contact point, with $\mathbf{P}_{F_c}(s_c)$ denoting the position along the backbone at arc length $s_c$ obtained from the static model. The circumferential offset term accounts for the $45^\circ$-spaced cable layout and is non-negligible under asymmetric tension states. Since $\mathbf{F}_c$ and $s_c$ are both unknown, they are estimated in two decoupled steps.

\subsubsection{Two-Step Decoupled Estimation}

\paragraph{Step~1 -- Contact force from sensor measurements.}
At the proximal end, each cable $C_a$ enters the instrument at a known fixed direction $\mathbf{d}_{C_a,0}$ determined by the capstan geometry and independent of robot configuration. The strain gauges directly measure the
proximal cable tensions $T_{C_a,0}$. Force equilibrium then yields the contact force as:
\begin{equation}
    \mathbf{F}_c = \mathbf{F}_s
    - \sum_{a=1}^{8} T_{C_a,0}\,\mathbf{d}_{C_a,0}.
    \label{eq:fc_direct}
\end{equation}
%This computation is non-iterative and requires no knowledge of the robot shape.

\paragraph{Step~2 -- Contact location from moment and length residuals.}
With $\mathbf{F}_c$ known, the static model (Algorithm~\ref{Algorithm_F2S}) is invoked with $\mathbf{F}_c$ and a current estimate of $s_c$ to obtain the robot shape $\{\mathbf{P}_{b,i},\mathbf{P}_{t,i}\}$ and the cable channel positions $\mathbf{P}_{C_a,j}$. The moment residual is:
\begin{equation}
    \mathbf{M}_\mathrm{res} = \mathbf{M}_s -
    \mathbf{M}_\mathrm{th},
    \label{eq:Mres}
\end{equation}
and the model-predicted cable length change is:
\begin{equation}
    \Delta L_{C_a}^{\mathrm{m}}(s_c)
    = \sum_{j} \bigl\|\mathbf{P}_{C_a,j+1}
    - \mathbf{P}_{C_a,j}\bigr\|
    - L_{C_a}^{0},
    \label{eq:dL_model}
\end{equation}
where $L_{C_a}^{0}$ is the cable length in the undeformed configuration. The contact location is recovered by minimizing a normalized cost function:
{\small
\begin{equation}
    \hat{s}_c = \arg\min_{s_c\in[0,L]}
    \frac{\bigl\|\mathbf{M}_\mathrm{res}\bigr\|^2}{M_{\max}^2}
    + \frac{\displaystyle\sum_{a=1}^{8}
    \bigl(\Delta L_{C_a}^{\mathrm{m}}(s_c)
    - \Delta L_{C_a}\bigr)^2}{L_{\max}^2},
    \label{eq:sc_opt}
\end{equation}}
where $M_{\max}$ and $L_{\max}$ are the measurement ranges of the torque channels and the cable displacement encoders, respectively, normalizing the two terms to a common dimensionless scale. Since $\mathbf{M}_\mathrm{th}$ and $\Delta L_{C_a}^{\mathrm{m}}$ both depend on $s_c$ through Algorithm~\ref{Algorithm_F2S}, Step~2 is embedded in a fixed-point iteration: $\mathbf{F}_c$ is computed once from Eq.~(\ref{eq:fc_direct}) and held fixed, while $s_c$ is updated by solving Eq.~(\ref{eq:sc_opt}) at each iteration until $|s_c^{(n+1)} - s_c^{(n)}| < e_{\mathrm{limit}}$. The optimization is solved using MATLAB's built-in solver.

\section{Results}

\begin{table}[t]
	\centering
	\caption{Structural parameters of notched  continuum surgical robots}
    \resizebox{0.48\textwidth}{!}{	
	\begin{tabular}{cccccccccccc}
\hline
\multicolumn{9}{c}{Structural parameters of segment A.}\\
\hline
		 $D_o$  & $d_c$ & $w$ & $h$& $L^i_{c}$ & $N_c$ & $E$(GPa) & $E_m$(GPa) & $\mu$ \\ 
        3.5 & 2.6 & 0.6 & 0.7 & 0.9 & 5&47.05&0.08&0.23\\ 
        \hline 
\multicolumn{9}{c}{Structural parameters of segment B.}\\
\hline
		 $D_o$  & $d_c$ & $w$ & $h$& $L^i_{c}$ & $N_c$ & $E$(GPa) & $E_m$(GPa) & $\mu$ \\ 
        3.5 & 2.6 & 0.4 & 0.7 & 0.9 & 5&47.05&0.08&0.23\\ 
        \hline         
	\end{tabular}
    }
	\label{NCR}
    \vspace{-10 pt}
\end{table}

 \begin{figure*}[t]
\centering  
\includegraphics[width=180mm]{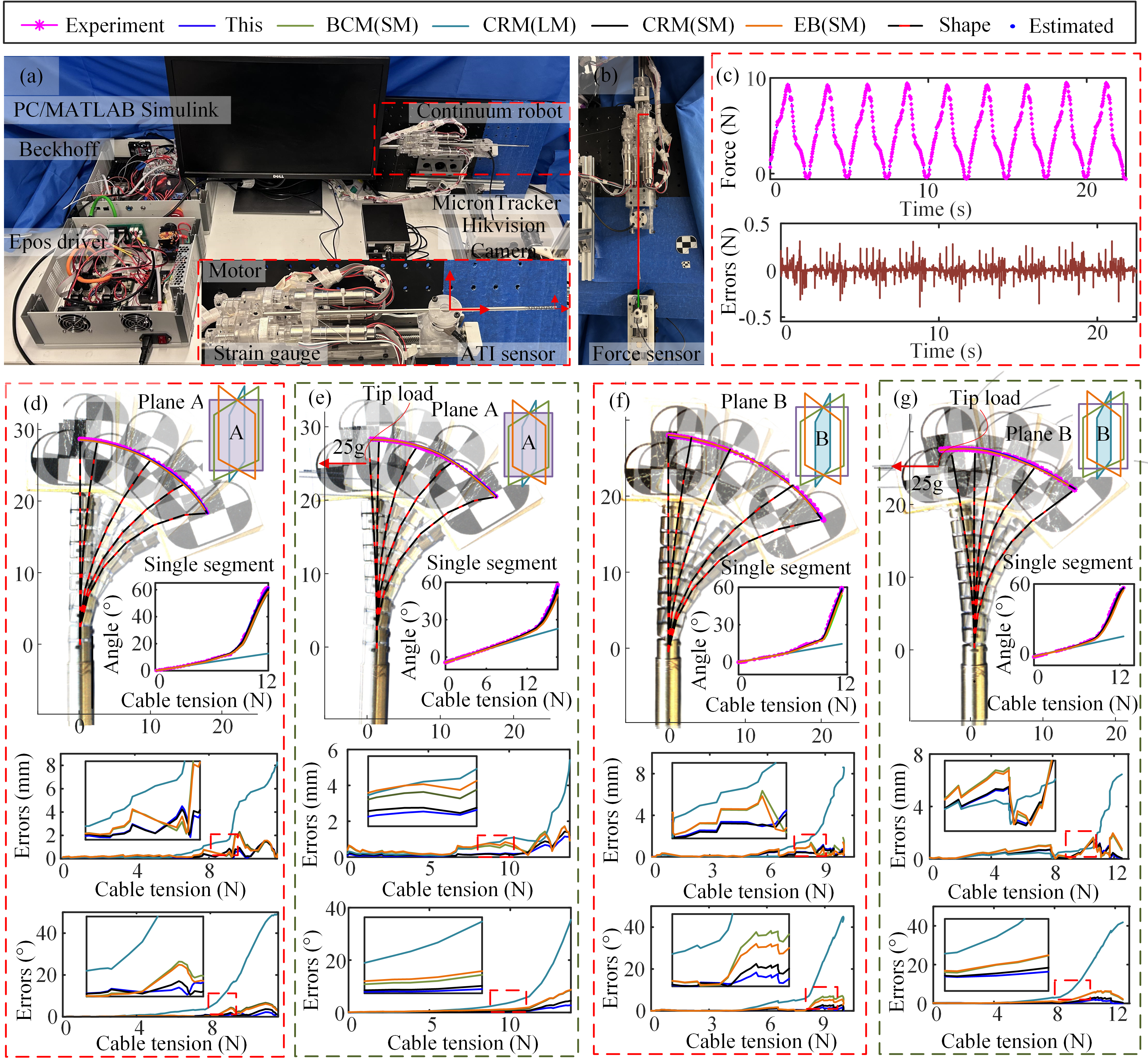}
\caption{Experimental setup and validation of the modeling framework for a single-segment continuum robot design. (a) Experimental setup for robot validation; (b) Experimental setup for validating cable tension sensing using strain gauges; (c) Error in force sensing using strain gauges; (d) Validation of bending in the plane A under no-load conditions; (e) Validation of bending in the plane A under a 25g load; (f) Validation of bending in the plane B under no-load conditions; (g) Validation of bending in the plane B under a 25g load.}
\label{Ex1}   
\vspace{-10 pt}
\end{figure*}
\subsection{Device}

The apparatus shown in Fig.~\ref{Ex1}(a) was used to verify the effectiveness of the constructed robot system. Two markers were fixed to the base and end effector of the robot's continuum manipulator, respectively. A binocular camera (MicronTracker, Claron Technology, Canada) was used to acquire the robot's motion trajectory information, which was transmitted to MATLAB Simulink via a communication interface. Data from the torque sensor was acquired via a data acquisition card and transmitted to MATLAB Simulink. The robot's drive data was transmitted to MATLAB/Simulink via TCP/IP communication.

Another camera (CS060, Hikrobot, China) was used to acquire the robot's shape information, which was saved using the official MVS software.
The validation was performed using two notched continuum robots (NCRs), one with only one segment (segment A), and the other with two segments (segment A and segment B), which were spaced 45 degrees apart along the axial direction. The robot's structure is shown in Fig.~\ref{CR}, and its parameters are shown in Table.~\ref{NCR}.

The contact force perception experiment was conducted using the same robotic platform. A calibrated weight was attached at the robot tip to generate a known external force for evaluating contact force estimation. Since the contact occurred at the distal tip, the ground-truth contact location was predefined as the tip position.

\subsection{Tension Calibration}
The device shown in Fig.~\ref{Ex1}(b) was used to calibrate the relationship between strain gauges and tension. A tension sensor (DYZ-100, Dayang, China) was fixed to one side. A cable tip with a rubber band was directly connected to the tension sensor. The cable was periodically moved by pulling a capstan to obtain data on the periodic changes. The information from the tension sensor and the strain gauge was converted into voltage signals by a transmitter and acquired by a data acquisition card (EL3164, Beckhoff, Germany). 

The drive cable reciprocates within a range of 0-10 mm, thereby applying a driving force to the cable. The relationship between the collected strain gauge data and tension sensor data is obtained using the Neural Network Toolbox in MATLAB. Nine sets of data were collected for verification; the maximum error in sensing tension using strain gauges did not exceed 0.4 N, and the average error was 0.12 N. 

The calibrated tension estimation model was subsequently applied to both single- and dual-segment continuum robot configurations for shape and contact perception experiments.
 
\subsection{Comparison with SOTA Models for Shape Perception in a Single-Segment Continuum Robot}

\begin{table*}[ht]
	\centering
\caption{Comparison results with SOTA models.}
\resizebox{\textwidth}{!}{	
\begin{tabular}{ccccclc|ccccccl}
\hline
\multicolumn{14}{c}{Case 1. Tip position prediction error in a single-segment continuum robot without tip load/with 25 g tip load in the (X-Z plane).}\\
\hline
\multirow{2}{*}{Load} &\multirow{2}{*}{Model} & \multicolumn{2}{c}{Angle}& \multicolumn{2}{c}{Position}&\multirow{2}{*} {Speed} &\multirow{2}{*}{Load}&\multirow{2}{*}{Model} & \multicolumn{2}{c}{Angle}& \multicolumn{2}{c}{Position}&\multirow{2}{*} {Speed} \\ 
&& {Mean Errors}& {Max Errors}& {Mean Errors}& {Max Errors}& & & &{Mean Errors}& {Max Errors}& {Mean Errors}& {Max Errors}&  \\ \hline
\multirow{6}{*}{0 g}&This& 1.12$\degree$& 2.22$\degree$& 0.34 mm&1.50 mm& 326 Hz&\multirow{6}{*}{25 g}&This& 1.42$\degree$& 2.48$\degree$& 0.36 mm&1.35 mm& 218 Hz\\
&This(NG)& 1.12$\degree$& 2.22$\degree$& 0.34 mm&1.50 mm& 24 Hz& &This(NG)& 1.42$\degree$& 2.48$\degree$& 0.36 mm&1.35 mm& 29 Hz\\
&BCM(SM)& 1.88$\degree$& 5.27$\degree$& 0.53 mm&2.15 mm& 108 Hz& &BCM(SM)& 2.82$\degree$& 7.06$\degree$& 0.62 mm&2.01 mm& 94 Hz\\
&CRM(SM)& 1.54$\degree$& 2.67$\degree$& 0.32 mm&1.61 mm& 11 Hz& &CRM(SM)& 1.24$\degree$& 2.92$\degree$& 0.38 mm&1.70 mm& 13 Hz\\
&CRM& 13.71$\degree$& 49.38$\degree$& 2.26 mm&7.62 mm& 17 Hz& &CRM& 11.02$\degree$& 43.28$\degree$& 1.61 mm&5.74 mm& 16 Hz\\
&EB(SM)& 1.86$\degree$& 5.15$\degree$& 0.55 mm&2.12 mm& $<$ 0.01 Hz&&EB(SM)& 2.91$\degree$& 7.12$\degree$& 0.63 mm&2.42 mm& $<$ 0.01 Hz\\
\hline
\multicolumn{14}{c}{Case 2. Tip position prediction error in a single-segment continuum robot in the (Y-Z plane).}\\
\hline
\multirow{2}{*}{Load} &\multirow{2}{*}{Model} & \multicolumn{2}{c}{Angle}& \multicolumn{2}{c}{Position}&\multirow{2}{*} {Speed} &\multirow{2}{*}{Load}&\multirow{2}{*}{Model} & \multicolumn{2}{c}{Angle}& \multicolumn{2}{c}{Position}&\multirow{2}{*} {Speed} \\ 
&& {Mean Errors}& {Max Errors}& {Mean Errors}& {Max Errors}& & & &{Mean Errors}& {Max Errors}& {Mean Errors}& {Max Errors}&  \\
\hline
\multirow{6}{*}{0 g}&This& 1.22$\degree$& 3.21$\degree$& 0.30 mm&1.16 mm& 261 Hz&\multirow{6}{*}{25 g}&This& 1.34$\degree$& 2.91$\degree$& 0.50 mm&1.42 mm& 213 Hz\\
&BCM(SM)& 2.43$\degree$& 6.43$\degree$& 0.42 mm&1.84 mm& 119 Hz& &BCM(SM)& 2.65$\degree$& 6.29$\degree$& 0.60 mm&1.84 mm& 87 Hz\\
&CRM(SM)& 1.38$\degree$& 3.69$\degree$& 0.24 mm&1.04 mm& 27 Hz& &CRM(SM)& 1.18$\degree$& 3.77$\degree$& 0.46 mm&1.40 mm& 19 Hz\\
&CRM& 12.51$\degree$& 45.02$\degree$& 2.18 mm&8.60 mm& 31 Hz& &CRM& 14.2$\degree$& 41.82$\degree$& 1.48 mm&6.44 mm& 17 Hz\\
&EB(SM)& 2.50$\degree$& 5.51$\degree$& 0.38 mm&1.40 mm& $<$ 0.01 Hz&&EB(SM)& 2.80$\degree$& 6.45$\degree$& 0.58 mm&1.98 mm& $<$ 0.01 Hz\\
\hline
\multicolumn{14}{c}{Case 3. Tip position prediction error in a two-segment continuum robot.}\\
\hline
\multirow{2}{*}{Load} &\multirow{2}{*}{Model} & \multicolumn{2}{c}{Angle}& \multicolumn{2}{c}{Position}&\multirow{2}{*} {Speed} &\multirow{2}{*}{Load}&\multirow{2}{*}{Model} & \multicolumn{2}{c}{Angle}& \multicolumn{2}{c}{Position}&\multirow{2}{*} {Speed} \\ 
&& {Mean Errors}& {Max Errors}& {Mean Errors}& {Max Errors}& & & &{Mean Errors}& {Max Errors}& {Mean Errors}& {Max Errors}&  \\
\hline
\multirow{6}{*}{0 g}&This& 1.38$\degree$& 2.45$\degree$&0.48 mm&1.38 mm& 296 Hz
&\multirow{6}{*}{25 g}&This& 1.76$\degree$& 2.58$\degree$& 0.62 mm&1.64 mm& 207 Hz\\
&BCM(SM)& 2.08$\degree$& 3.75$\degree$& 0.82 mm&2.26 mm& 121 Hz&
&BCM(SM)& 2.11$\degree$& 3.78$\degree$& 0.83 mm&1.74 mm& 98 Hz\\
&CRM(SM)& 1.30$\degree$& 2.14$\degree$& 0.54 mm&1.80 mm& 29 Hz& 
&CRM(SM)& 1.62$\degree$& 2.41$\degree$& 0.58 mm&2.06 mm& 15 Hz \\
&CRM& 5.97$\degree$& 23.19$\degree$& 2.64 mm&7.84 mm& 43 Hz& 
&CRM& 5.62$\degree$& 17.53$\degree$& 1.44 mm&5.18 mm& 23 Hz\\
&EB(SM)& 2.21$\degree$& 3.87$\degree$& 0.66 mm&1.84 mm& $<$ 0.01 Hz&
&EB(SM)& 2.02$\degree$& 2.90$\degree$& 0.84 mm&2.11 mm& $<$ 0.01 Hz\\
\hline
\end{tabular}
}
\label{comp}
\vspace{-10 pt}
\end{table*}

 \begin{figure*}[t]
\centering  
\includegraphics[width=180mm]{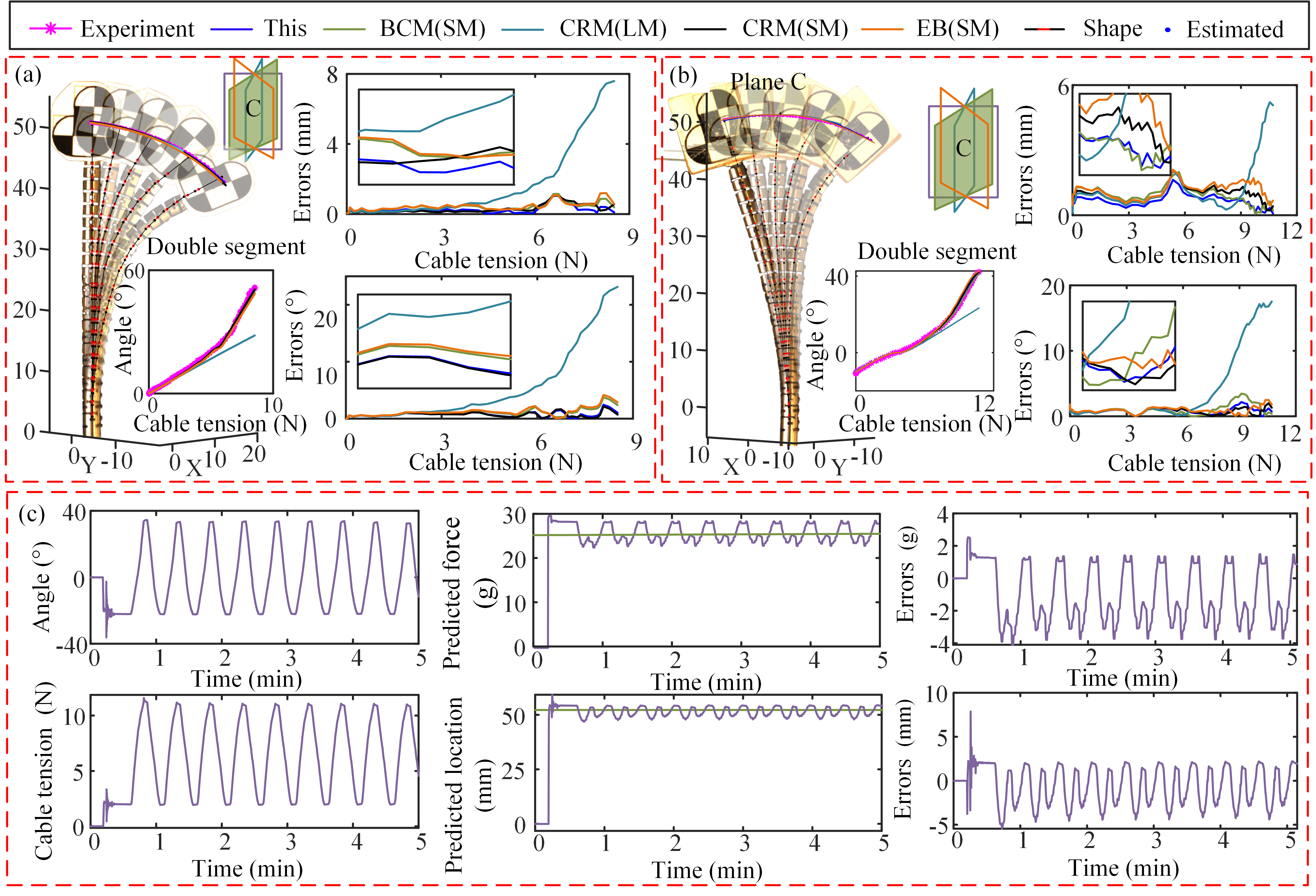}
\caption{Validation of the Modeling Framework and Contact Force Sensing in the Design of a Two-Segment Continuum Robot. (a) Bending validation in the 45°–135° plane under no-load conditions; (b) Bending validation in the 45°–135° plane under a 25 g load; (c) Validation of the modeling framework and contact perception in a two-segment continuum robot.
}
\label{Ex2}   
\vspace{-15 pt}
\end{figure*}

To facilitate the description of cable orientations, four possible bending planes were defined and denoted as Plane A--D, corresponding to cable orientations of 0°/180°, 90°/$-$90°, 45°/$-$135°, and $-$45°/135°, respectively. Here, two representative bending configurations (Plane A and Plane B) were experimentally evaluated under no-load and 25 g tip-load conditions.
The cables moved at a constant speed within 0-3.2 mm, causing the robot to bend. The pitch-plane bending results are displayed in the rotated horizontal frame, as shown in Fig.~\ref{Ex1}(d-g). The proposed framework was compared against four baseline models: a beam constraint model (BCM(SM)), a Cosserat rod model with material nonlinearity (CRM(SM)), a Cosserat rod model without material nonlinearity (CRM), and a Bernoulli--Euler beam model (EB(SM)). Note that material nonlinearity was not considered in the original CRM formulation; it is incorporated here for a fair comparison. A variant of the proposed model with group parallel processing disabled, denoted This(NG), is also included to isolate the effect of the parallel computation strategy.

Results are summarized in Table~\ref{comp}. In the plane A without load (Case~1), the proposed model achieves a mean tip position error of 0.34 mm (max 1.50 mm) at 326\,Hz, comparable in accuracy to CRM(SM) (0.32 mm at 19 Hz) while running at nearly 30$\times$ the speed. All other baselines show larger errors. Under a 25 g load (Case~1, 25g), accuracy is well maintained at 0.36 mm mean (max 1.35 mm) at 218 Hz. In the plane B (Case~2), mean position errors of 0.30 mm (no load, 261 Hz) and 0.50 mm (25 g load, 213 Hz) are achieved, again at substantially higher speed than comparable baselines. Across all single-segment cases, disabling group parallel processing reduces speed from over 200 Hz to 24-29 Hz.

\subsection{Model Validation and Contact Perception in a Two-Segment Continuum Robot}

For the two-segment configuration (Case~3, Plane C), the proposed model achieves mean tip position errors of 0.48\,mm (max 1.38\,mm) at 296\,Hz under no load and 0.62\,mm (max 1.64\,mm) at 207\,Hz under a 25\,g load. Here, the three-dimensional robot configuration is projected onto Plane C to facilitate comparison between the reconstructed shape and camera images. Due to inter-segment coupling and the 45° axial offset between segments, the actual robot configuration is not strictly constrained within Plane C, but exhibits a spatial deformation close to this reference plane. The prediction accuracy is comparable to CRM(SM), which yields mean position errors of 0.54 mm and 0.58 mm, respectively. In terms of computational efficiency, however, the proposed model achieves approximately 10.2$\times$ and 13.8$\times$ higher update rates than CRM(SM) under the two conditions, while also outperforming BCM(SM), CRM, and EB(SM) in mean position accuracy. Full results are given in Table~\ref{comp}.

The perception framework was further evaluated on the two-segment robot through a reciprocating contact experiment, as shown in Fig.~\ref{Ex2}(c). The estimated contact force and location were compared with their corresponding reference values. The mean contact force estimation error was 1.89 g, with a maximum error of 4.09 g, the mean contact location estimation error was 1.93 mm, with a maximum error of 7.87 mm.

\section{Discussion}

This paper presents a method for measuring cable tension in a capstan-driven continuum surgical robot by incorporating flexible deformation elements fitted with strain gauges into the motor mounts—a design that achieves tension sensing without encroaching upon the already compact space within the capstan assembly. This design preserves the inherent structural compactness and rapid-exchange capabilities of the capstan-driven system while providing critical inputs for shape and contact force perception. Experimental results demonstrate that, within the 0–9.5 N tension range, the average error is 0.12 N, with a maximum error of 0.4 N. More importantly, the availability of cable tension, together with proximal force/torque measurements and the proposed static model, enables shape and contact force perception to be realized in the capstan-driven system. Therefore, the primary significance of the proposed sensing design lies not in outperforming dedicated inline force sensors in measurement accuracy, but in enabling perception within a compact capstan-driven architecture where direct integration of such sensors is difficult.

The current prototype utilizes 3D-printed structural components and a process involving the attachment of strain gauges using cyanoacrylate adhesive. This approach is adequate for supporting experiments during the proof-of-concept phase, and the data obtained validate the feasibility of the proposed sensing methodology. However, this fabrication process exhibits significant limitations regarding stability and consistency: batch-to-batch variations in the mechanical properties of 3D-printed parts, adhesive aging and creep, and positional deviations of the strain gauges resulting from manual application all introduce systemic errors. Experiments revealed significant zero-point drift between trials, necessitating recalibration prior to each run—a requirement that, to some extent, constrains the system's "plug-and-play" capability. Consequently, the current fabrication process is suitable only for laboratory-based proof-of-concept studies; before proceeding to long-term research or clinical application, dedicated optimization of the manufacturing process to enhance stability is required. This paper focuses primarily on validating the feasibility of the sensing architecture; specific process improvements—such as the use of precision-machined mounts, high-temperature-curing adhesives, or even integrated MEMS fabrication techniques—are reserved for future work.

For the two-segment configuration, the measured robot shape was projected onto the corresponding reference bending plane for visualization and comparison with camera observations. Due to the 45° axial offset between segments, the actual robot configuration exhibits three-dimensional deformation rather than being strictly confined to the reference plane. Although the proposed model considers inter-segment coupling and spatial cable interactions, the increased structural complexity and accumulated uncertainties in the multi-segment configuration introduce additional modeling challenges, resulting in a slight increase in prediction error compared with the single-segment case.
The mean tip position error increases modestly from the single-segment case (0.48 mm vs. 0.34 mm under no load). The contact localization accuracy achieves a maximum error of 7.87 mm, which is mainly affected by the observability of deformation-related signals. When the applied contact force is small, the resulting cable length variation and moment change become less significant, leading to reduced sensitivity in contact position estimation. This limitation is also observed in model-based contact localization methods, where localization performance depends on the magnitude of induced deformation~\cite{zhang2026cable}. Meanwhile, the contact force estimation results demonstrate that the proposed tension sensing scheme provides usable force information for contact perception without requiring sensing elements along the continuum body.
Extending the framework to three or more segments would not increase the dimensionality of the optimization problem, since $s_c$ remains a scalar unknown; however, model accuracy and convergence under more complex multi-segment interactions warrant further investigation.

\section{Conclusion}

This paper presents an integrated actuation--perception design framework for capstan-driven continuum surgical robots, addressing the challenge of obtaining cable tension information within a compact capstan assembly. By integrating compliant tension-sensing structures into the motor mount, designing compact spatial cable routing, developing a multibody static model, and integrating the proximal perception strategy, the proposed approach enables shape and contact force perception while preserving the compactness and rapid-exchange capability of capstan-driven systems.

Experimental results demonstrate that the proposed tension sensing module achieves an average error of 0.12 N over a 0--9.5 N range. The developed static model achieves real-time computation with update rates exceeding 200 Hz while maintaining accuracy comparable to Cosserat beam models. The proposed perception framework further enables contact force and location estimation in continuum robots, demonstrating the feasibility of integrating sensing capability into compact capstan-driven surgical instruments.

Future work will focus on improving the long-term stability of the strain gauge fabrication process, developing more efficient solvers for multi-segment configurations, and conducting in vivo validation.

\bibliographystyle{ieeetr}
\bibliography{Reference}
\vspace{-110 pt}

\begin{IEEEbiography}[{\includegraphics[width=1in,height=1.20in,clip,keepaspectratio]{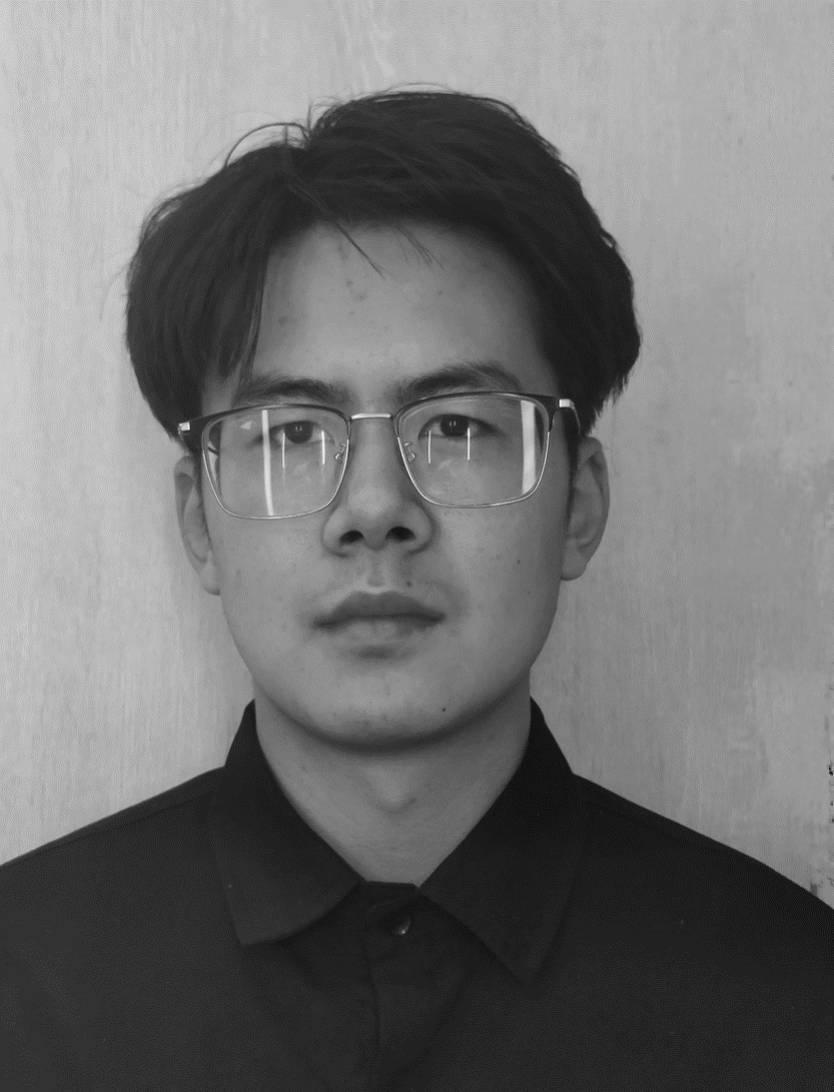}}]{Gang Zhang}
received his bachelor's and master's degrees from the Department of Mechanical Engineering of Shandong University in 2021 and 2024. He is studying for his doctor's degree in the Department of Mechanical and Automation Engineering of The Chinese University of Hong Kong currently. His research interests include robotic modeling, control, and perception of continuum robots for surgery.
\end{IEEEbiography}
%%%%%%%%%%%%%%%%%%%%%%%%%%%%%
\vspace{-110 pt}

\begin{IEEEbiography}[{\includegraphics[width=1in,height=1.20in,clip,keepaspectratio]{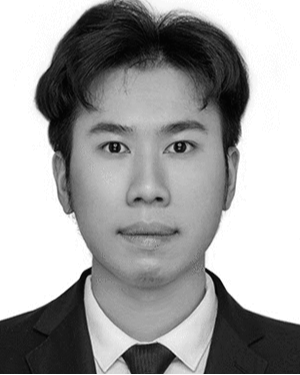}}]{Yufu Qiu} received the B.Eng. degree in automation from the Guangdong University of Technology, Guangzhou, China, in 2021. He is currently working toward the Ph.D. degree in mechanical and automation engineering with the Department of Mechanical and Automation Engineering, The Chinese University of Hong Kong, Hong Kong, China.
His research interests include the design and control of surgical robotic systems.
\end{IEEEbiography}
%%%%%%%%%%%%%%%%%%%%%%%%%%%%%
\vspace{-110 pt}

\begin{IEEEbiography}[{\includegraphics[width=1in,height=1.20in,clip,keepaspectratio]{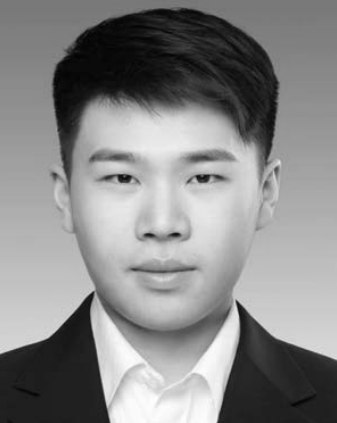}}]{JunYan Yan}
(Member, IEEE) received his Bachelor of Engineering degree in Mechanical Design, Manufacturing and Automation from Shandong University in 2019. He received his PhD degree in Mechanical and Automation Engineering from the Chinese University of Hong Kong in 2024. He is currently a postdoctoral fellow in the Department of Mechanical and Automation Engineering at the Chinese University of Hong Kong.
\end{IEEEbiography}
%%%%%%%%%%%%%%%%%%%%%%%%%%%%%
\vspace{-110 pt}

\begin{IEEEbiography}[{\includegraphics[width=1in,height=1.20in,clip,keepaspectratio]{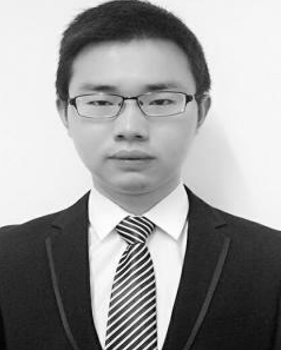}}]{Wenhui Zeng}
associate Professor at the School of Mechanical and Electronic Engineering, Wuhan University of Technology. Research focuses on flexible catheter robots, flexible medical surgical robots and precision machinery.
\end{IEEEbiography}
%%%%%%%%%%%%%%%%%%%%%%%%%%%%%
\vspace{-110 pt}

\begin{IEEEbiography}[{\includegraphics[width=1in,height=1.20in,clip,keepaspectratio]{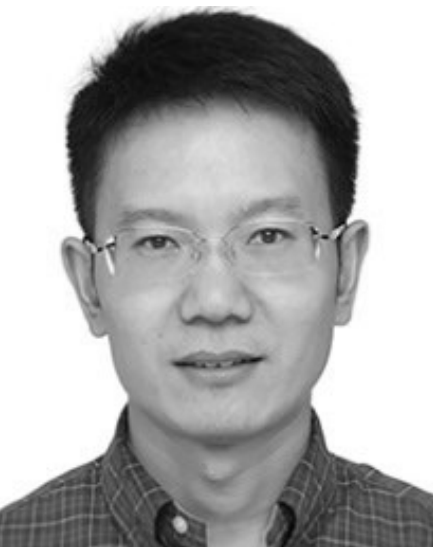}}]{Wenlong Lu} received the Ph.D. degree in precision engineering from EPSRC Centre for Advanced Metrology, University of Huddersfield, Huddersfield,U.K., in 2012. From 2016 to 2018, he worked as a Visiting Scholar with the University of Illinois at Urbana–Champaign, Urbana, IL, USA. Now he is a Professor with the School of Mechanical Science and Engineering, Huazhong University of Science and Technology, Wuhan, China. His research interests focused on advanced optical sensing technology, precision measurement technology and instrumentation, friction and wear.
\end{IEEEbiography}
%%%%%%%%%%%%%%%%%%%%%%%%%%%%%
\vspace{-110 pt}

\begin{IEEEbiography}[{\includegraphics[width=1in,height=1.20in,clip,keepaspectratio]{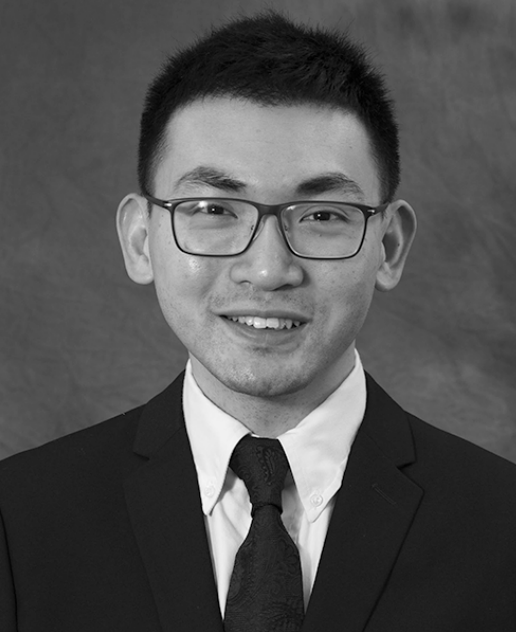}}]{Shing Shin Cheng}(Member, IEEE) received the B.S. degree in mechanical engineering from the Johns Hopkins University, USA, in 2013, and Ph.D. degree in robotics from the Georgia Institute of Technology, USA, in 2018. He is currently an Associate Professor in the Department of Mechanical and Automation Engineering, The Chinese University of Hong Kong, Hong Kong, China. His research interests include flexible surgical robotics, image-guided surgical systems, and robot modeling and control.
\end{IEEEbiography}
%%%%%%%%%%%%%%%%%%%%%%%%%%%%%
\vspace{-100 pt}

% that's all folks
\end{document}